\documentclass[10pt,twocolumn,letterpaper]{article}

\usepackage{cvpr}
\usepackage{times}
\usepackage{epsfig}
\usepackage{graphicx}
\usepackage{amsmath}
\usepackage{amssymb}

\usepackage[pagebackref=true,breaklinks=true,letterpaper=true,colorlinks,bookmarks=false]{hyperref}

\cvprfinalcopy 

\def\cvprPaperID{****} 

\ifcvprfinal\fi
\begin{document}

\title{Face Alignment by Local Deep Descriptor Regression}


\author{Amit Kumar\\
University of Maryland\\
College Park, MD 20742\\
{\tt\small akumar14@umd.edu}
\and
Rajeev Ranjan\\
University of Maryland\\
College Park, MD 20742\\
{\tt\small rranjan1@umd.edu}
\and
Vishal M. Patel\\
Rutgers University\\
New Brunswick, NJ 08901\\
{\tt\small vishal.m.patel@rutgers.edu}
\and
Rama Chellappa\\
University of Maryland\\
College Park, MD 20742\\
{\tt\small rama@umiacs.umd.edu}
}

\maketitle

\begin{abstract}
   We present an algorithm for extracting key-point descriptors using deep convolutional neural networks (CNN). Unlike many existing deep CNNs, our model computes local features around a given point in an image. We also present a face alignment algorithm based on regression using these local descriptors. The proposed method called \textit{Local Deep Descriptor Regression (LDDR)} is able to localize face landmarks of varying sizes, poses and occlusions with high accuracy. Deep Descriptors presented in this paper are able to uniquely and efficiently describe every pixel in the image and therefore can potentially replace traditional descriptors such as SIFT and HOG. Extensive evaluations on five publicly available unconstrained face alignment datasets show that our deep descriptor network is able to capture strong local features around a given landmark and performs significantly better than many competitive and state-of-the-art face alignment algorithms.
\end{abstract}

\section{Introduction}

Face alignment is a crucial component of applications such as face recognition, face verification and expression analysis. Most of the recent methods use discriminatory shape regression approach to estimate the face landmark positions. With their ability to utilize large amount of training data, and enforce shape constraints adaptively, regression-based methods have achieved state-of-the-art performance on various unconstrained face alignment datasets. However, the success of these methods is limited by the strength of the features they use. In previous works, the features used are either hand crafted ; for example SIFT was used as features in \cite{XiongD13}, or learned from a limited set of training samples \cite{DBLP:journals/ijcv/CaoWWS14,DBLP:conf/cvpr/RenCW014}.

\begin{figure}[t]
\label{overview_fig}
\begin{center}
\includegraphics[width = 0.4\textwidth]{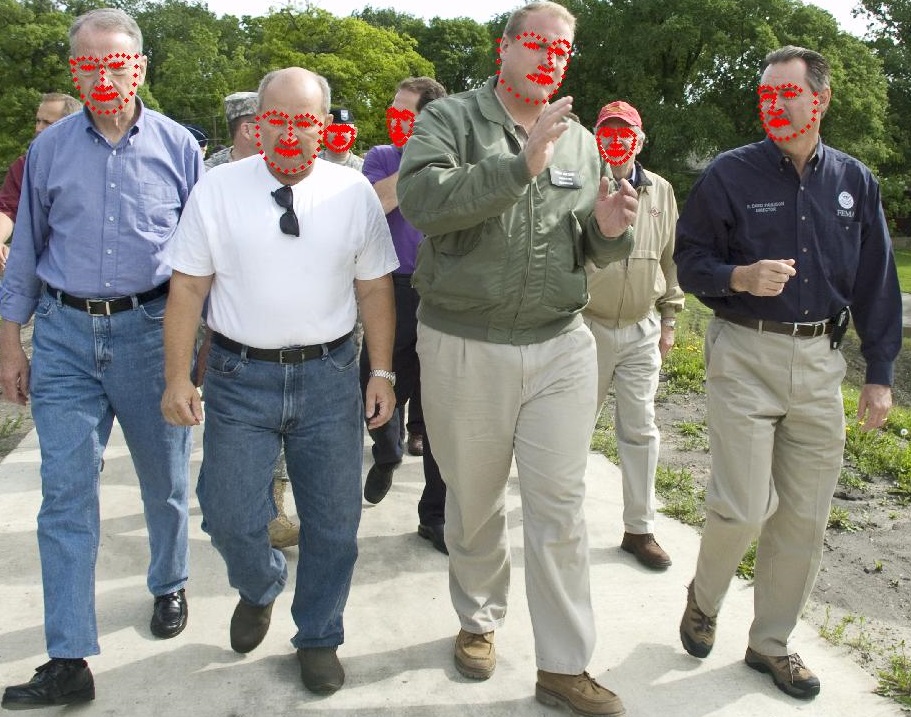}
\end{center}
\vskip-10pt
\caption{We present a deep descriptor-based regression approach for fiducial point extraction.  This figure shows fiducial points extracted on all the detected faces on an image from the IJB-A\cite{F._2015_CVPR} dataset using our method.}
\end{figure}

\begin{figure*}[t]
\begin{center}
\includegraphics[width =.8 \textwidth]{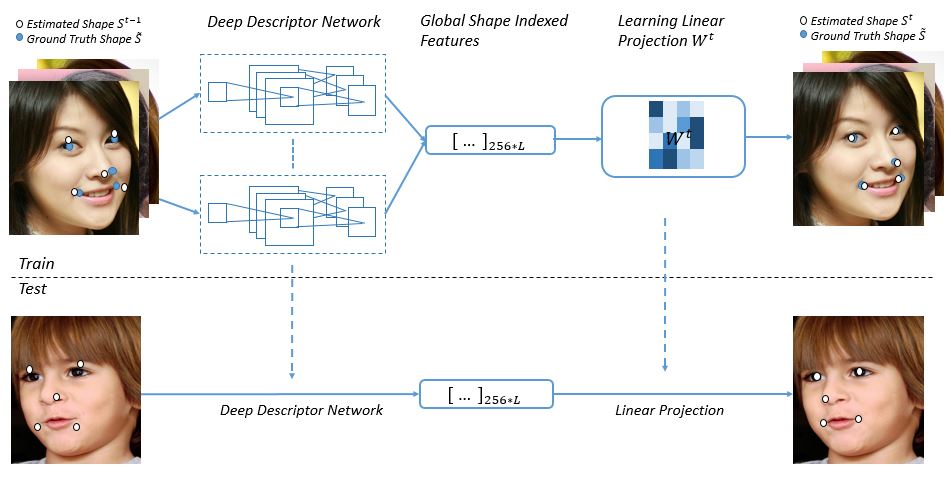}
\end{center}
\vskip-17pt
\caption{Overview of our method. During training, we extract \textit{deep descriptors} for each landmark  and concatenate them to form a shape-indexed feature vector. Given these features and target shape increments $\Delta S_{i}^t$, we learn the linear regression weights \textit{$W^t$}. During testing, deep descriptors are extracted around each point of the initialized mean shape. Intermediate shape is predicted using the regressor weights \textit{$W^t$}. This process is iterated to reach the final estimated shape.}\label{overview}
\end{figure*}

In recent years, features obtained using deep CNNs have yielded impressive results for various computer vision applications. They significantly outperform methods proposed earlier for the tasks of face detection and recognition. It has been shown in \cite{NIPS2012_4824} that a deep CNN pre-trained with a large generic dataset such as Imagenet \cite{ILSVRC15}, can be used as a meaningful feature extractor. Although these features are strong enough for reliable classification, they are global in nature. Hence, this approach may not be effective for problems such as face alignment where local features are desirable. To overcome this problem Overfeat \cite{sermanet-iclr-14}  uses predicted detection boundaries, but  lacks the needed pixel-based localization feature. \cite{Taylor:2010:CLS:1888212.1888225} and \cite{10.1109/TPAMI.2012.231} propose pixel-based localization, the former based on Restricted Boltzmann machine while the latter processes the image to determine a key-point descriptor.

In this paper, we solve the localization problem in existing deep CNNs by constructing a deep convolutional key-point descriptor model. We build a network which takes a small local image patch around a pixel as an input and produces a feature vector as the output. We claim that the proposed deep descriptor network can be used as a substitute for SIFT \cite{Lowe04distinctiveimage} descriptors in most vision problems. To support our claim, we apply the descriptor model for facial landmark detection. Local features calculated for a small rectangular patch around each estimated landmark position are used by a linear regressor to learn the shape increment during training, and predict the landmark positions at test time. Figure~\ref{overview_fig} shows several faces where our method is able to locate fiducial points on all the detected faces.  Overall, this paper makes the following contributions:
\begin{enumerate}
\item We construct a novel deep descriptor network to evaluate the local features for a given key-point.
\item We perform face alignment by applying linear regression to the deep descriptors evaluated for facial landmarks.
\end{enumerate}

 Section~\ref{prev_work} reviews a few related works.  Details of our deep descriptor-based face alignment method are given in Section~\ref{gen_inst}.  Section~\ref{experiments} provides the landmark localization results on five challenging datasets. Finally, Section~\ref{sec:conc} concludes the paper with a brief summary and discussion.

\section{Previous Work}
\label{prev_work}
The task of face alignment can be classified broadly into three categories depending on the approach.
\subsection{Model-based Approaches}
Model-based approaches learn a shape model during training and uses it to fit new faces during testing. The pioneering works of Cootes \textit{et al.} such as Active Appearance Models (AAM) \cite{AAM} and Active Shape Models (ASM) \cite{Cootes:1995:ASM:206543.206547} were built using PCA constraints on appearance and shape. In recent years many improvements over these models have been proposed in \cite{Matthews:2004:AAM:993451.996344,conf/eccv/LiangXWS08,Gross:2005:GVP:1709142.1709186,Gross:2006:AAM:1709247.1709267,Sauer11accurateregression,doi:10.5244/C.24.91}.  In \cite{Cristinacce06featuredetection}, Cristinacce and Cootes generalised the ASM model to a Constrained Local Model (CLM), in which every landmark has a shape constrained descriptor to capture the appearance. In \cite{Saragih:2011:DMF:1937966.1938021}, a more sophisticated local model and mean shift was used to achieve good results. However, these methods depend upon the goodness of the error cost function and how well it is optimised. For example, AAM estimates the shape by minimizing the texture residual. Recently Antonakos \textit{et al.} \cite{Antonakos_2015_CVPR} proposed method along similar lines by modeling the appearance of the object using multiple graph-based pairwise normal distributions (Gaussian Markov Random Field) between the
patches extracted from the regions. However, the learned models lack the power to capture complex face image variations in pose, expression and illumination. Also, they are sensitive to initialization due to gradient descent optimization, a critical step.

\subsection{Regression-based Approaches}
Since face alignment is naturally a regression problem there has been a plethora of regression-based approaches in recent years. These methods learn a regression model that directly maps image appearance to target output. But the performance of these methods depend on the robustness of local descriptors. Sun et al \cite{Sun:2013:DCN:2514950.2516090} proposed a cascade of carefully designed CNNs in which at each level outputs of multiple networks are fused for landmark estimation. Our work is different from \cite{Sun:2013:DCN:2514950.2516090}, in a way that we use a single CNN carefully designed to provide a unique key-point descriptor. Xiong \textit{et al.} \cite{XiongD13} predicts the increment in shape by applying linear regression on SIFT features. Burgos \textit{et al.}\cite{Welinder10p.:cascaded} proposed a cascade of T-regressors to estimate the pose in image sequence using pose-indexed features. Cao \textit{et al.} \cite{DBLP:journals/ijcv/CaoWWS14} sequentially learned a cascade of random fern regressors using pixel intensity difference as the feature and regresses the shape stage-wise over the learnt cascade. They performed regression on all parameters simultaneously, thus effectively exploiting the shape constraint. Following this, Sun \textit{et al.} \cite{DBLP:conf/cvpr/RenCW014} proposed cascaded regression using fern regressors and local binary features. Subsequently, Burgos \textit{et al.} \cite{10.1109/ICCV.2013.191} extended their work to face alignment with occlusion handling, enhanced shape indexed features and more robust initialization which they call as Robust cascaded pose regression (RCPR). Li \textit{et al.} \cite{Yan:2013:LCM:2586110.2586274} combined multiple final shapes from multiple initializations in a cascade regression manner using weights matrices learnt to combine these hypotheses accurately. Recently, Lee \textit{et al.} \cite{Lee_2015_CVPR} proposed a Gaussian Process Regression face alignment method based on the responses of the Gaussian filters around the patches extracted from the region adjacent to intermediate landmarks. Zhu \textit{et al.} \cite{Zhu_2015_CVPR} proposed a hierarchical face alignment , starting from a coarse shape estimate and refining it to reach the target landmark. Also Xiong {et al.} \cite{global2015xiong} proposed global supervised descent method where they consider directly optimizing over the landmarks independent of any shape model.   

\subsection{Part-based Deformable Models}
Part-based deformable models perform alignment by maximizing the posterior likelihood of part locations given an input image \textit{I}. The models vary in the optimization techniques or the shape priors used. In \cite{conf/iccv/SaragihLC09} Saragih \textit{et al.} used a method similar to mean shift to optimize the posterior likelihood. Recently, Saragih \cite{DBLP:conf/cvpr/Saragih11} developed a sample specific prior which significantly improves over the original PCA prior in ASM , CLM and AAM. Zhu and Ramanan \cite{AFW_dataset_CVPR2012} used a part-based model for face detection, pose estimation and landmark localization assuming the face shape to be a tree structure. Asthana \textit{et al.} \cite{Asthana:2013:RDR:2514950.2516059}  combined discriminative response map fitting with CLM, which learns a dictionary of probability response maps based on local features and adopts linear regression-based fitting in the CLM framework. 

\section{Regression of Deep Descriptors}
\label{gen_inst}

The proposed method for facial landmark detection, called \textbf{Local Deep Descriptor Regression} (LDDR), consists of two modules. The first module generates local features for each estimated facial landmark points using the deep descriptor framework. These features are concatenated together to form a global shape-indexed feature. The second module is a linear regressor which learns the relationship between the shape feature and the corresponding shape increment during training. The process is repeated stage-by-stage in a cascaded fashion. Figure \ref{overview} shows the overview of our method.

\subsection{Deep Descriptor Construction}
\label{deep_descriptor}

 In order to construct a deep CNN descriptor, we start with the Alexnet \cite{NIPS2012_4824} network. We use the publicly available network weights trained on the Imagenet \cite{ILSVRC15} data using Caffe \cite{jia2014caffe}, that are distributed with RCNN \cite{girshick14CVPR} . However, this particular CNN cannot be used directly as a key-point descriptor because of the following limitations. Firstly, the CNN requires a fixed input image size of $224 \times 224$ pixels which is too large to be considered for the patch size around the key-point. Secondly, a single activation unit at the fifth convolutional layer ($conv_5$) has a highly overlapping receptive field of size $195 \times 195$ pixels, which makes localization difficult. As a result, two pixel points in close vicinity cannot be distinguished from one another.

On further analysis of the first problem, we found that a CNN requires fixed size input only because of its fully-connected layers. A convolutional layer can process any input as long as it is larger than the convolutional kernel. On the other hand, a fully connected layer needs a fixed size input as its output dimension is predetermined. To resolve this issue, we remove the last max pooling layer ($pool_5$) and all the subsequent fully-connected layers ($fc_6$, $fc_7$, $fc_8$, and softmax) from the network. The CNN output is, therefore, computed by the $conv_5$ layer containing 256 feature channels.
\begin{figure*}[t!]
\begin{center}
\includegraphics[width =.85\textwidth]{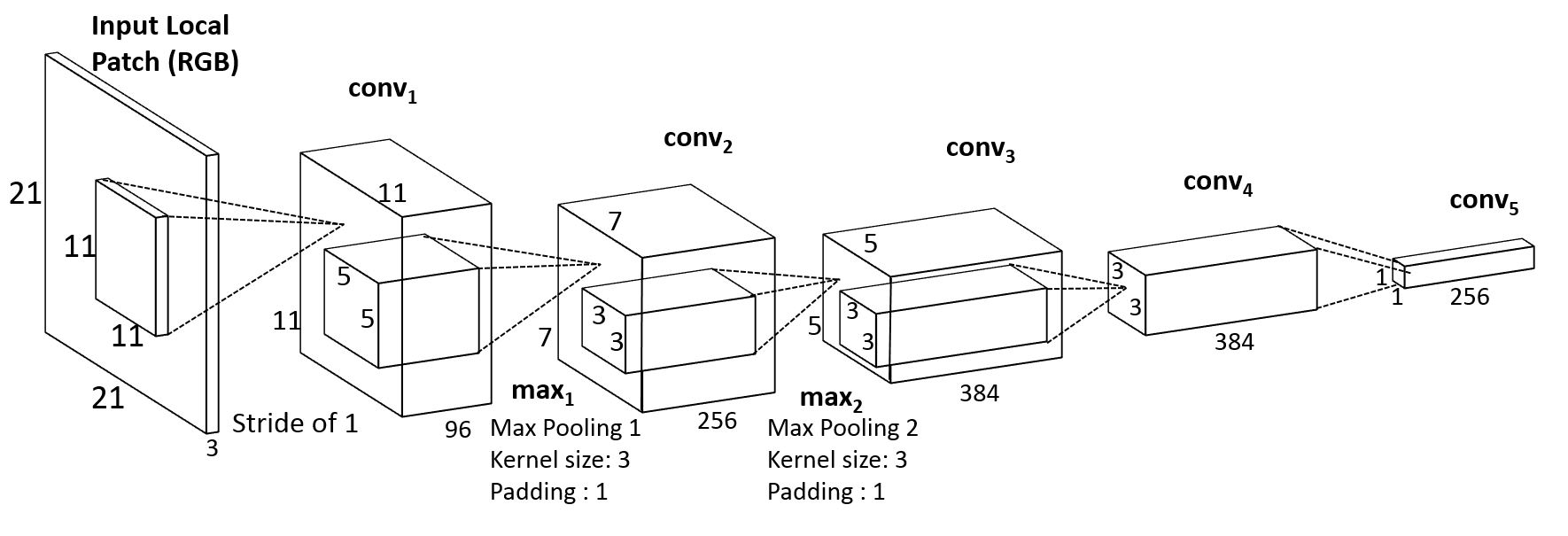}
\end{center}
\caption{Architecture of the proposed Deep Descriptor Network. The height and width represents the dimensions of each feature map, whereas the depth denotes the number of features maps for a given layer. The number of strides for each layer is restricted to 1.}
\label{architecture}
\end{figure*}
Analyzing the second problem, we find that a major contributor for the large size of receptive field is the inter-layer subsampling operation, which is implemented in the form of strides in the convolutional as well as max pooling layers. They are deployed mainly to reduce the number of parameters and feature computation time, which are not required for a key-point descriptor since the small patch input will drastically bring down the convolution time anyway. Hence, strides in all the existing layers are set to $1$. Also, padding from all the convolutional layers are removed as they contribute very little to describing a key-point. Instead, we apply a single pixel padding in the max pooling layer to further reduce the size of the receptive field without disturbing the output. With these architectural changes, the receptive field size is reduced to $21 \times 21$ pixels which is good enough for the size of a local patch surrounding a key-point. The final network structure obtained for the deep descriptor is shown in Figure \ref{architecture}.  With the input size as small as the receptive field, single pixel feature maps are obtained at the $conv_5$ layer forming a 256 dimensional output vector.

The proposed deep descriptor satisfies the essential properties of being a key-point descriptor. It is position independent, as it depends only on the image patch relative to the point. It is robust to small geometric transformations because of the max pooling operation in CNN. The normalization operation after each convolutional layer makes it robust to illumination variations. Since the network weights are trained using fixed sized inputs, the descriptor works best when the input images are scaled to the same size prior to key-point extraction, thus reducing the dependency on scale. Hence it can be used as a generic keypoint descriptor in many computer vision applications. Additionally, for domain specific problems, the model weights can be fine-tuned before evaluating the features. For the application of face alignment, we fine-tune the model weights using face images from the FDDB \cite{fddbTech} dataset. Fine-tuning was done for the face detection task, which classifies the input as face or non-face. The procedure adopted is similar to the method described in \cite{girshick14CVPR}. During fine-tuning, the network learns features specific to face parts which is a crucial part in our work. As a result, the activations at the fifth convolutional layer become more discriminatory to local face patches such as eyes, nose, lips, etc. The other advantage of fine-tuning is that the same network weights can be used for both face detection as well as face alignment. Once the network is fine-tuned, the test image just goes through a forward pass to generate CNN features, which are then fed to a simple linear regression method to generate incremental shapes.

\subsection{Computing Shape Indexed Features}

Given an initial mean shape containing $L$ landmarks, we compute the $256$ dimensional deep descriptor $\phi_l^t$ for each landmark $l \in {1,2,....L}$ at a given stage $t$. A global shape indexed feature is composed by concatenating the set of deep descriptors, i.e., $\Phi{\textsuperscript{\textit{t}}}=[\phi_1^t,\phi_1^t,...,\phi_L^t]$ , which is subsequently used to learn the ground truth shape increment, as explained in section \ref{learn_regressor}.

We adopt a coarse to fine regression approach. It is important in face alignment that the features used to describe the landmark points are local. To predict the offset $ \Delta\textit{s}$ of a single landmark, we extract the deep descriptors from a local region of size \textit{r}. It has been shown in \cite{DBLP:conf/cvpr/RenCW014} that the optimal size is almost linear to the standard deviation of individual shape increment $\Delta\textit{s}$. Since, we want $\Delta\textit{s}$ to decrease sharply at every stage, we need to choose the size of the local patch region around the landmark accordingly. Following \cite{DBLP:conf/cvpr/RenCW014}, we keep the patch size for deep descriptor larger in the first stage and decrease it linearly in subsequent stages. With this modification, the deep descriptor is bound to generate higher dimensional output for the initial stages. Additional structural modification is needed for uniform output dimension, which limits us to consider only four stages of regression. The patch sizes normalized by face rectangle are taken to be ${0.4, 0.3, 0.2, 0.1}$ for respective stages. Since the face is resized to $224 \times 224$ pixels (the input face size used for fine-tuning), the actual patch sizes correspond approximately to ${92, 68, 42, 21}$. Moreover, variable amounts of strides are added to $conv_1$, $max_1$, $conv_2$ and $max_2$ layers for each stage as listed in Table \ref{strides}. The network for the last stage remains unchanged as its input patch size matches the requirement for our deep descriptor network. This ensures a consistent output dimension of $256$ at each stage and for every landmark. In addition to just removing the fully connected layers, our network has reduced the amount of subsampling/stride for different regression stages as shown in Table \ref{strides}.

\begin{table}[htp!]
\begin{center}
\resizebox{\linewidth}{!}{%
\begin{tabular}{|p{1.1cm}|p{1.6cm}|p{0.7cm}|p{.7cm}|p{.7cm}|p{.7cm}|}
\hline
 \centering Stage 1 &  \centering Input Size (pixels) &  \centering conv1 &  \centering max1 & \centering conv2 &  max2\\
\hline\hline
Stage 1        &\centering $92 \times 92$       &\centering 4    &\centering 2    &\centering 1     & 1 \\
Stage 2          &\centering $68 \times 68$          &\centering 3    &\centering 2     &\centering 1     & 1 \\
Stage 3           &\centering $42 \times 42$              &\centering 2    &\centering 1     &\centering 1     & 2 \\
Stage 4            & \centering $21 \times 21$                 &\centering 1    &\centering 1     &\centering 1    & 1 \\
\hline
\end{tabular}}
\vskip 4pt
\caption{Input size and the number of strides in conv1, max1, conv2 and max2 layers for 4 stages of regression.} \label{strides}
\end{center}
\end{table}

\subsection{Learning Global Regression \textit{W\textsuperscript{t}}}
\label{learn_regressor}

In this section we introduce our basic shape regression methodology for the face alignment problem. Unlike \cite{DBLP:journals/ijcv/CaoWWS14} and \cite{DBLP:conf/cvpr/RenCW014} which have two level cascaded regression framework, we perform a single global regression at each stage.
Given a face image \textit{I} and initial shape $\textit{S\textsuperscript{0}}$, the regressor computes the shape increment $\Delta S$ from the deep descriptors and updates the face shape using  (\ref{update_eq})
\begin{equation}
\label{update_eq}
 \textit{$S^{t} $} = \textit{$S^{t-1}$} + \textit{$W^t$}\Phi^{t}(\textit{I},\textit{$S^{t-1}$}).
\end{equation}
After extracting the deep descriptors, we concatenate them to a form a global shape-indexed feature $\Phi{\textsuperscript{\textit{t}}}=[\phi_1^t,\phi_1^t,...,\phi_L^t]$. Our aim is to learn a global linear projection
\textit{W\textsuperscript{t}} by minimizing the following objective function:
\begin{equation}
\label{objective_function}
 \underset {\textit{$W^t$}} \min \sum_{i=1}^{N} \lVert \Delta \tilde{S^\textit{t}_\textit{i}}-\textit{W\textsuperscript{t}} \Phi \textsuperscript{\textit{t}}(\textit{$I_{i}$},\textit{$S_i^{t-1} $})\rVert_2^2 + \lambda\lVert \textit{$W^t$}\rVert_2^2,
\end{equation}
where the first term is the regression target and the second term is a regularization of $\textit{W\textsuperscript{t}} $ in $L2$ sense. $ \lambda $ controls the strength of regularization. Regularization here plays a major role due to the high dimensionality of the shape-indexed feature. In the experiments, the dimensionality of features for $68$ landmarks points could be as high as 17K+. Without regularization there could be substantial amount of over-fitting.
For implementing regression, we use L2 regularized L2-loss support vector regression using the LIBLINEAR \cite{REF08a} package. Since the objective function is quadratic in $W\textsuperscript{t}$, we can always reach a global minimum.

\subsection{Incorporating Shape Constraint} As mentioned in \cite{DBLP:journals/ijcv/CaoWWS14}, the shape constraint is preserved by learning a vector regressor and explicitly minimizing the shape alignment error as in (\ref{objective_function}). Since each shape is updated in an additive manner, and each shape increment is a linear combination of certain training shapes, the final shape is modeled as a linear combination of the initial shape \textit{$S^0$} and all training shapes:
\begin{equation}
\textit{S} = \textit{$S^0$} + \sum_{\textit{i}=1}^{\textit{N}} \textit{$w_i$}\hat{\textit{$S_i$}}.
\end{equation}
Hence, as long as the initial shape satisfies the shape constraint, the regressed final shape is bound to lie in the linear subspace constructed by all the training shapes. As a matter of fact all the intermediate shapes also satisfy the shape constraint, since they are constructed in a similar fashion.

\section{Experiments}
\label{experiments}
There are several landmark annotated datasets publicly available. However, we choose the most recent and challenging ones. These are Helen \cite{Le:2012:IFF:2403072.2403124}, LFPW \cite{Belhumeur:2011:LPF:2191740.2192193}, AFW \cite{AFW_dataset_CVPR2012} and IBUG \cite{6595977}.   In addition to these, we evaluate the performance of our method on a recently introduced IARPA Janus Benchmark A (IJB-A) dataset \cite{F._2015_CVPR}.   These datasets present different variations in face shape, appearance and pose and are described in the following subsections. To maintain consistency in the experiments, we perform face alignment using MultiPie \cite{Gross:2010:MUL:1746745.1747071} 68 point markup format.
\begin{figure*}[!htp]
 \centering
\includegraphics[width=6.1cm]{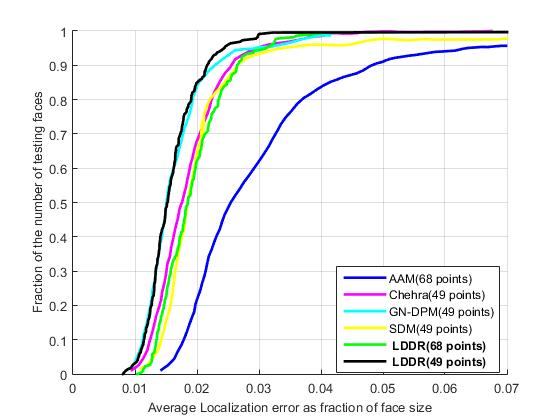}\hskip.1pt \includegraphics[width=6.1cm]{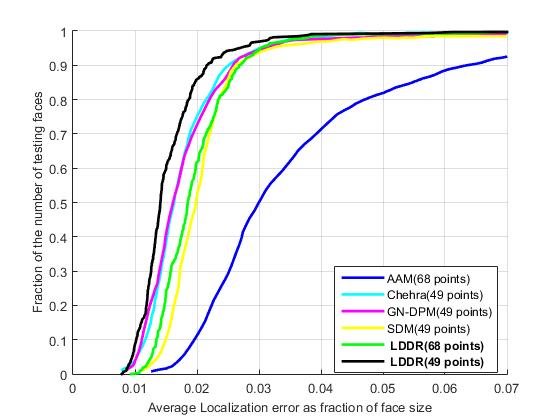}\\
(a)\hskip 200pt (b)\\
\includegraphics[width=6.1cm]{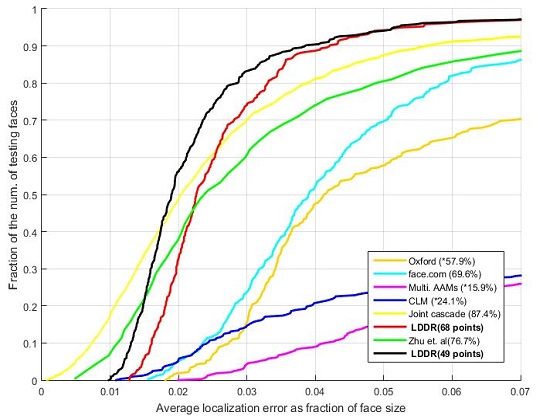}\hskip.1pt \includegraphics[width=6.1cm]{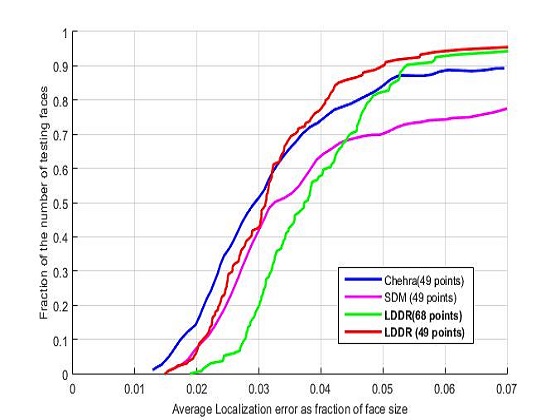}\\
(c) \hskip200pt (d)
\vskip 4pt
\caption{Average pt-pt error (normalized by face size) vs fraction of images in (a) LFPW, (b) Helen, (c) AFW and  (d) iBUG.}
\label{fig:Errors}
\end{figure*}

\subsection{Datasets}
\textbf{LFPW} \cite{Belhumeur:2011:LPF:2191740.2192193} is one of the widely used datasets to benchmark the face alignment tasks. It consists of $811$ training and $220$ testing images. The dataset contains unconstrained images from the internet which have large variations in pose, illumination and expression. Since some of the image links mentioned in the dataset are invalid, we downloaded the LFPW images from the ibug \cite{6595977} website which has accumulated all valid images and their $68$ point annotations. \par
\textbf{Helen} \cite{Le:2012:IFF:2403072.2403124} dataset has $2300$ high resolution web images, each one marked with $194$ landmark points. To be consistent with the $68$ point markup in our experiments, we downloaded this dataset from the ibug website which provides the $68$ point annotations along with this dataset. \par
\textbf{AFW} has annotated faces in the wild dataset created by Zhu and Ramanan \cite{AFW_dataset_CVPR2012}. It consists of $205$ in-the-wild-faces with varying illumination, pose, attributes and expressions. It was originally annotated with $6$ landmark points. However, we perform our experiment on the AFW dataset provided on ibug website, as it contains $68$ points annotated ground truth helping us to maintain consistency in the experiments.\par
\textbf{IBUG} is a challenging subset of $135$ images taken from the \textit{300-W} \cite{6595977} dataset. \textit{300-W} contains IBUG and images from existing datasets LFPW, Helen, AFW and XM2VTS \cite{Messer99xm2vtsdb:the}. It inherently follows the $68$ point annotation format. \par
\textbf{IJB-A} \cite{F._2015_CVPR} dataset is the recently released face verification dataset. The dataset is annotated with 3 key-points on the faces (two eyes and nose base). The dataset contains images and videos from 500
subjects collected from online media. In total, there are 67,183 faces of which 13,741 are from images and
the remaining are from videos. The locations of all faces in the IJB-A dataset were manually annotated by human annotators. The subjects were captured so that the dataset contains wide geographic distribution. The challenge comes through the wide diversity in pose, illumination and resolution. 

\par
\textbf{Training and testing}: We evaluated the performance of our method on these challenging datasets.   First, we performed training and testing on the LFPW and Helen datasets taking only their own training and testing sets. Using this model we test on AFW dataset. In order to evaluate on the IBUG dataset, we generated our own cumulative training set consisting of 3148 images taken from the LFPW, Helen and AFW datasets. This is done since AFW has more pose variations compared to LFPW and Helen. To test on IJBA-A dataset we use the same model. \par

\textbf{Evaluation Metric:} Following the standards of \cite{DBLP:journals/ijcv/CaoWWS14}, \cite{Belhumeur:2011:LPF:2191740.2192193}, we computed the average error for all landmarks in an image normalized by the inter-pupil distance. For each dataset, the mean error evaluated over all the images is reported. In the following sub-section, we compare our LDDR algorithm against existing state-of-the-art methods and validate our results.
Since the IJB-A dataset has only three annotated points, the interoccular distance error was normalized by the distance between nose tip and the midpoint of the eye centers.

\subsection{Comparison with state-of-the-art Methods}

During training we augment the data to improve the generalization ability. A single training sample is translated to multiple samples by flipping all the images and then randomly rotating them. Then initial shapes are also randomly assigned. Our method has only one fitting parameter \textit{i.e.} number of stages of regression, which following the principles of \cite{DBLP:conf/cvpr/RenCW014}, \cite{DBLP:journals/ijcv/CaoWWS14} has been set to 4 in our case.    We compare our results with those reported in \cite{DBLP:journals/ijcv/CaoWWS14}, \cite{DBLP:conf/cvpr/RenCW014}, \cite{10.1109/ICCV.2013.191}, \cite{Asthana:2013:RDR:2514950.2516059}, \cite{6909635}.

\begin{table}[thp!]
\begin{center}
\resizebox{\linewidth}{!}{%
\begin{tabular}{|p{4.1cm}|p{1.5cm}|p{1.5cm}|}
\hline
\centering {\it Method} & \centering {\it 68-pts} &  {\it 49-pts}\\
\hline\hline
\centering Zhu \it{et al.} \cite{AFW_dataset_CVPR2012}   & \centering 8.29         & 7.78  \\
\centering DRMF \cite{Asthana:2013:RDR:2514950.2516059}         & \centering 6.57         & -      \\
\centering RCPR \cite{10.1109/ICCV.2013.191}         & \centering 6.56         & 5.48  \\
\centering SDM \cite{XiongD13}          & \centering 5.67         & 4.47    \\
\centering GN-DPM \cite{6909635}       & \centering 5.92         & 4.43  \\
\centering CFAN \cite{CFAN}       & \centering 5.44         & -    \\
\centering CFSS \cite{Zhu_2015_CVPR}       & \centering 4.87         & 3.78    \\
\hline
\centering {\bf LDDR}   & \centering {\bf 4.67}   & {\bf 2.38}\\
\hline
\end{tabular}}
\caption{Averaged error comparison of different methods on the LFPW dataset.} 
\label{tbl:LFPW}
\end{center}
\end{table}

\begin{figure*}[htp!]
 \centering
\includegraphics[width=2.2cm]{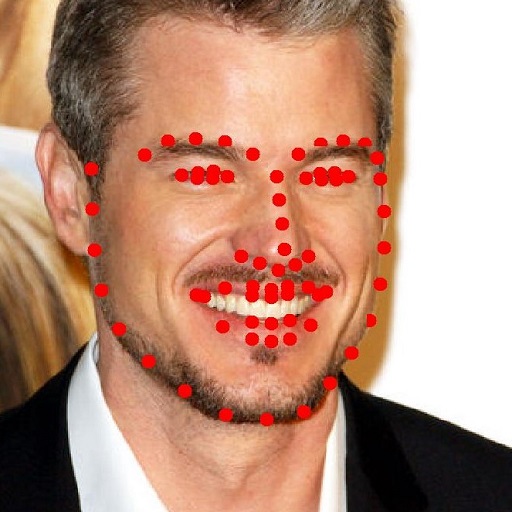}\includegraphics[width=2.2cm]{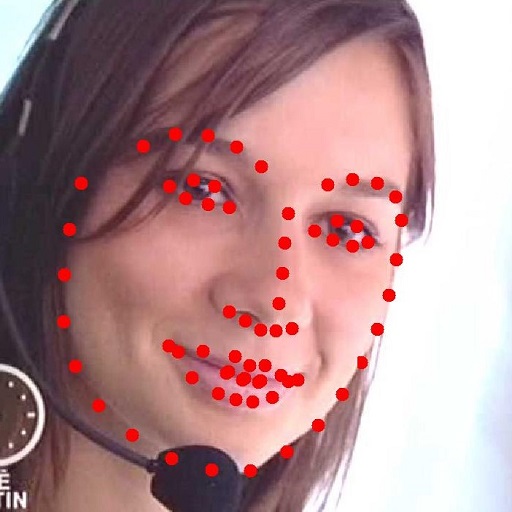}\hskip.1pt\includegraphics[width=2.2cm]{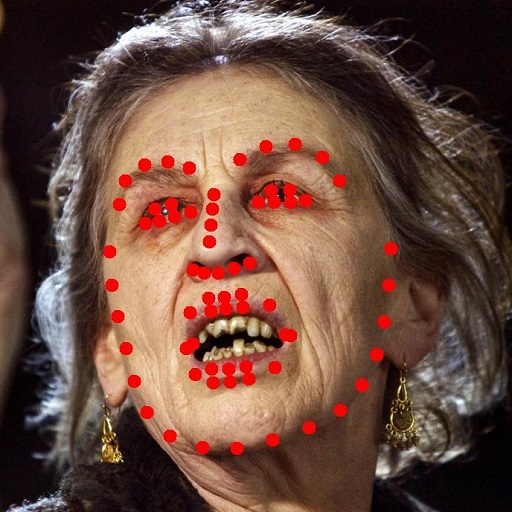}\hskip.4pt\includegraphics[width=2.2cm]{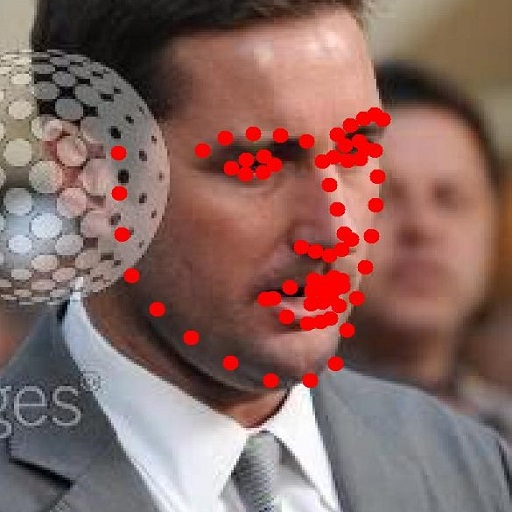}
\hskip.4pt\includegraphics[width=2.2cm]{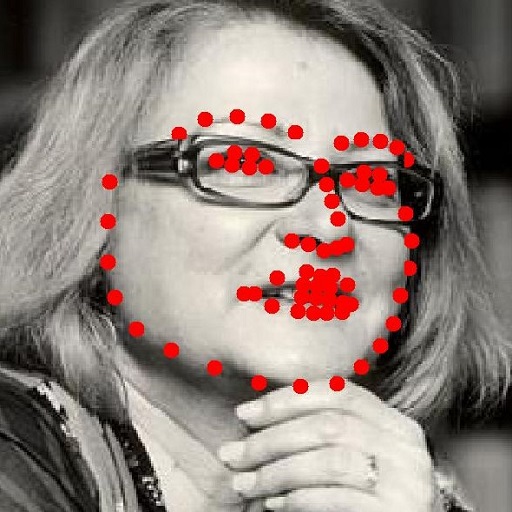}\hskip.4pt\includegraphics[width=2.2cm]{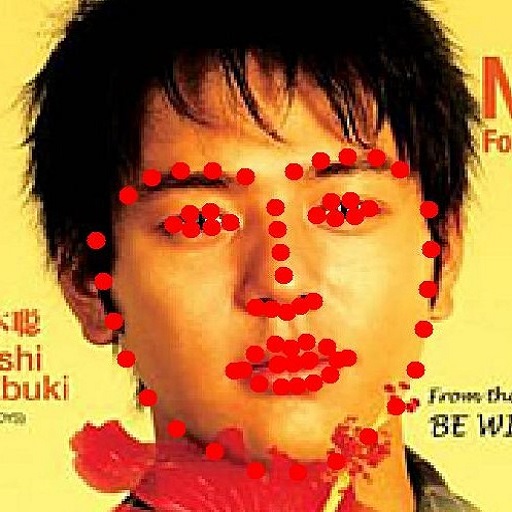}\\
\includegraphics[width=2.2cm]{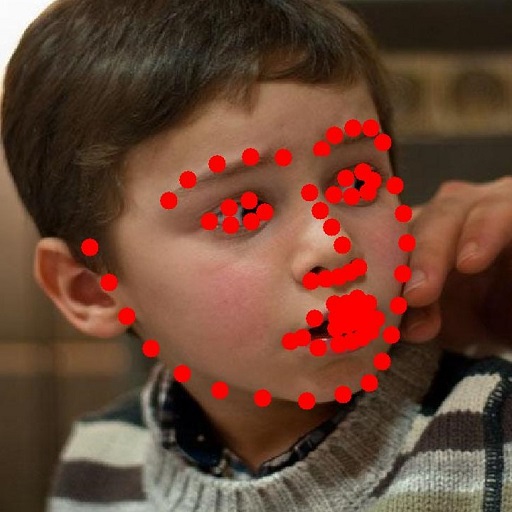}\hskip.2pt\includegraphics[width=2.2cm]{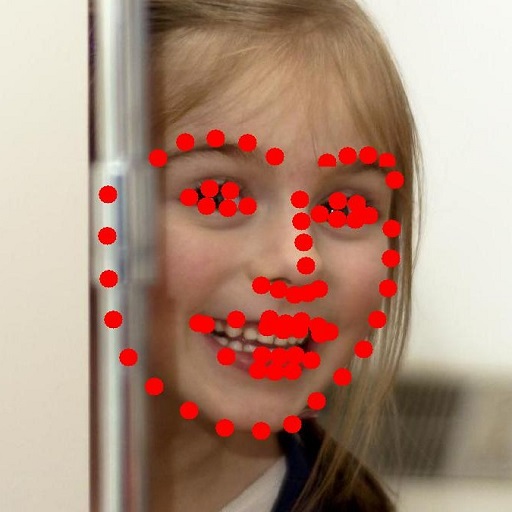}\hskip.2pt\includegraphics[width=2.2cm]{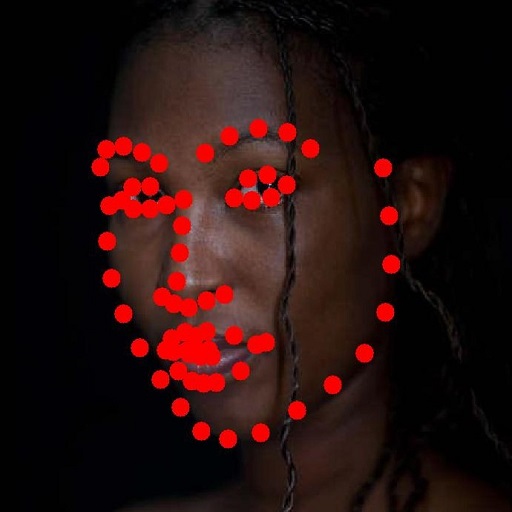}\hskip.2pt\includegraphics[width=2.2cm]{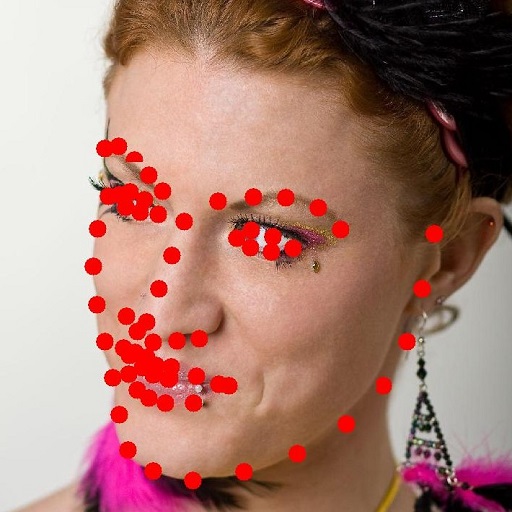}
\hskip.2pt\includegraphics[width=2.2cm]{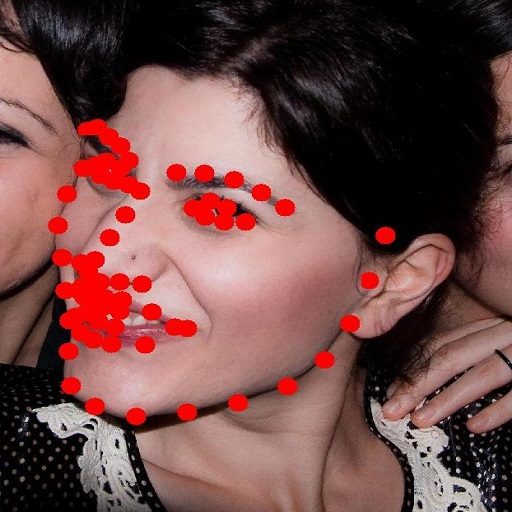}\hskip.2pt\includegraphics[width=2.2cm]{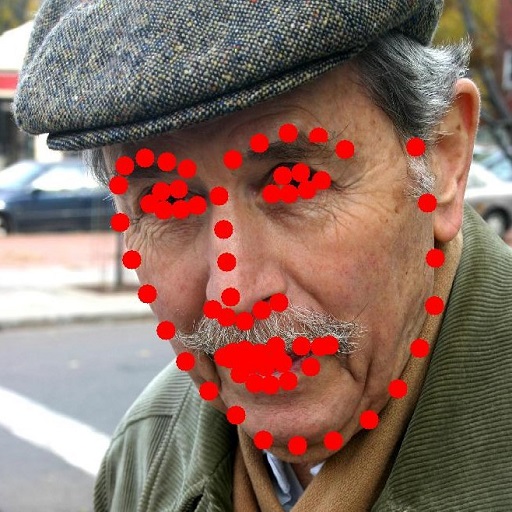}\\
 \includegraphics[width=2.2cm]{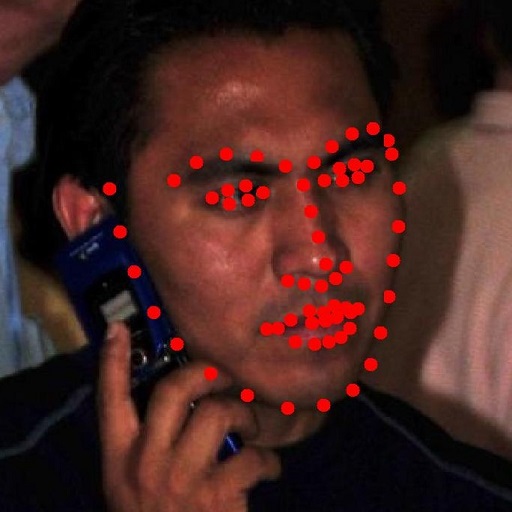}\hskip.2pt\includegraphics[width=2.2cm]{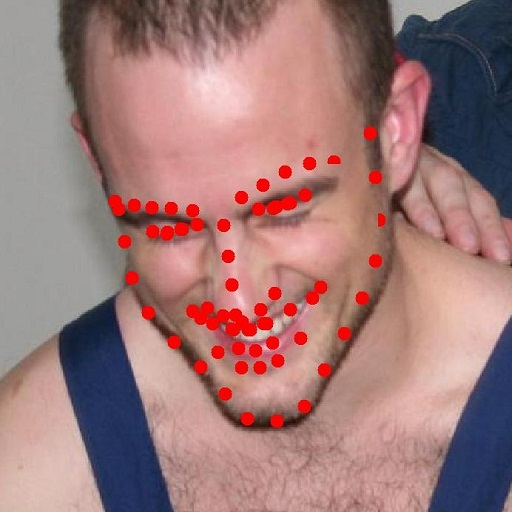}\hskip.2pt\includegraphics[width=2.2cm]{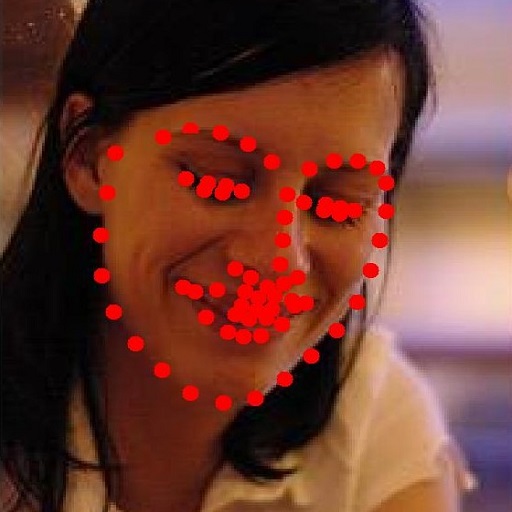}\hskip.2pt\includegraphics[width=2.2cm]{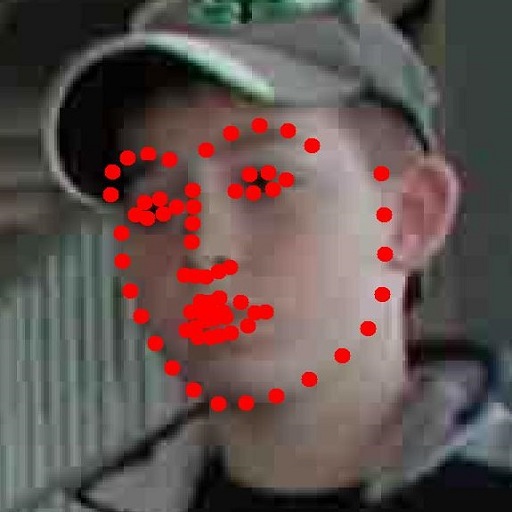}
 \hskip.2pt\includegraphics[width=2.2cm]{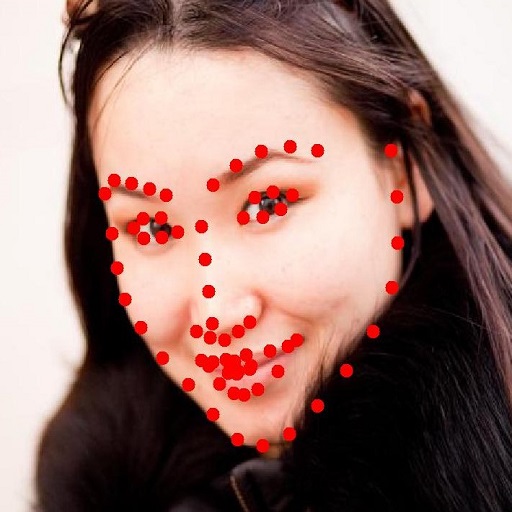}\hskip.2pt\includegraphics[width=2.2cm]{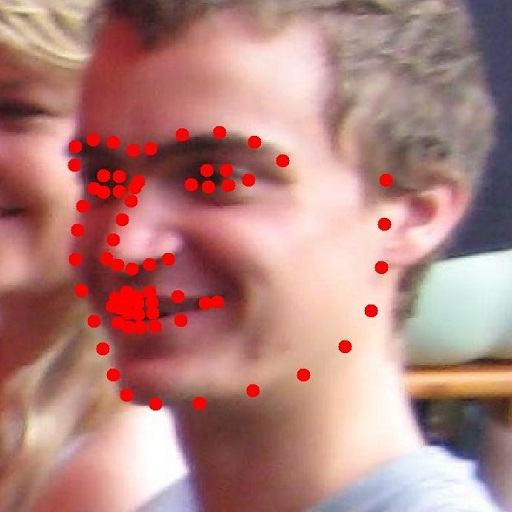}\\
 \includegraphics[width=2.2cm]{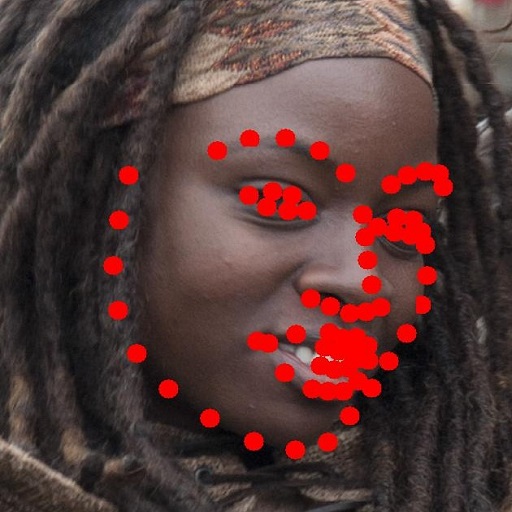}\hskip.2pt\includegraphics[width=2.2cm]{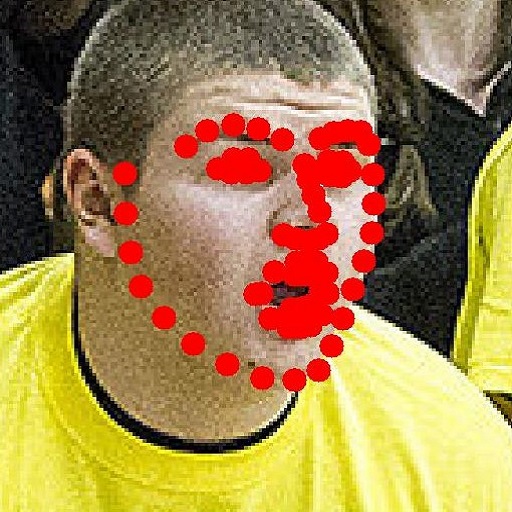}\hskip.2pt\includegraphics[width=2.2cm]{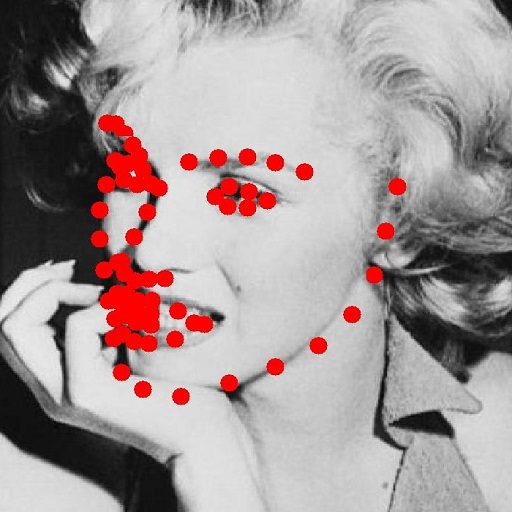}\hskip.2pt\includegraphics[width=2.2cm]{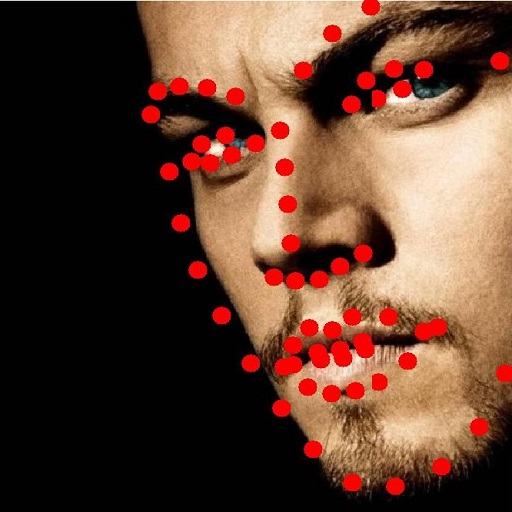}
 \hskip.2pt\includegraphics[width=2.2cm]{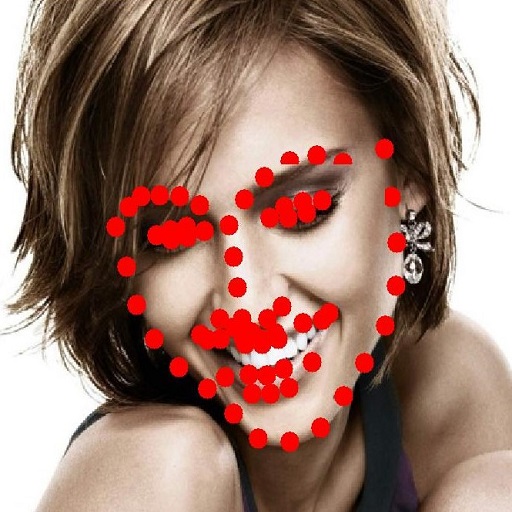}\hskip.2pt\includegraphics[width=2.2cm]{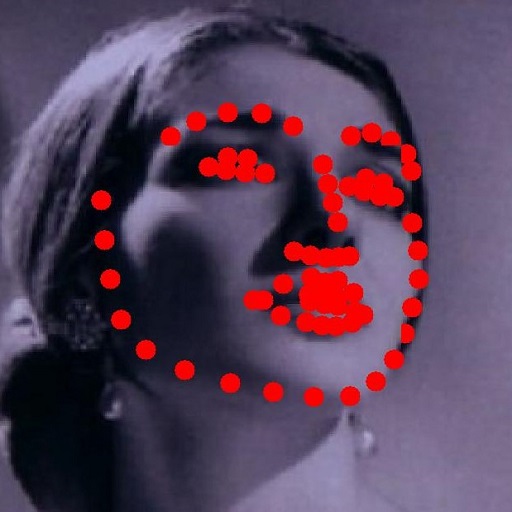}\\\includegraphics[width=2.2cm]{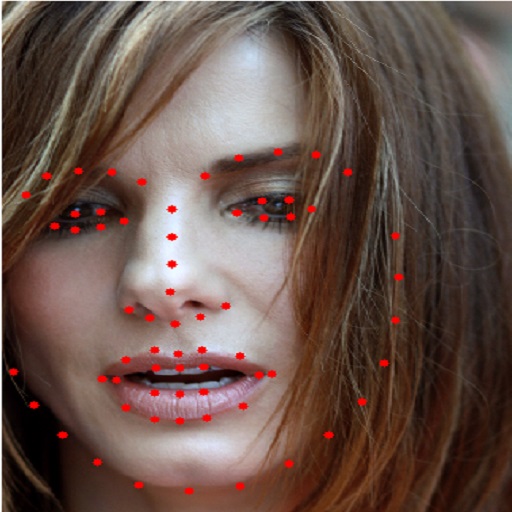}\hskip.2pt\includegraphics[width=2.2cm]{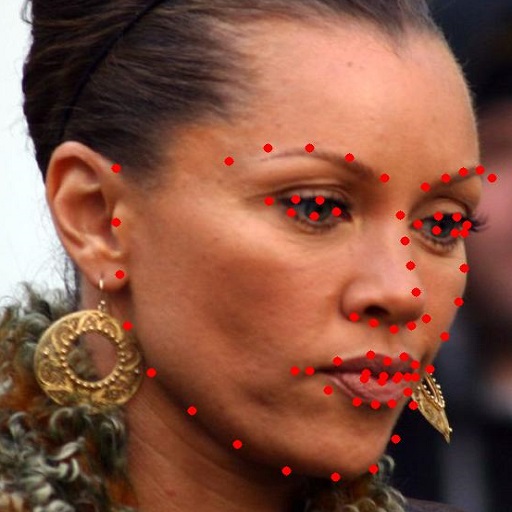}\hskip.2pt\includegraphics[width=2.2cm]{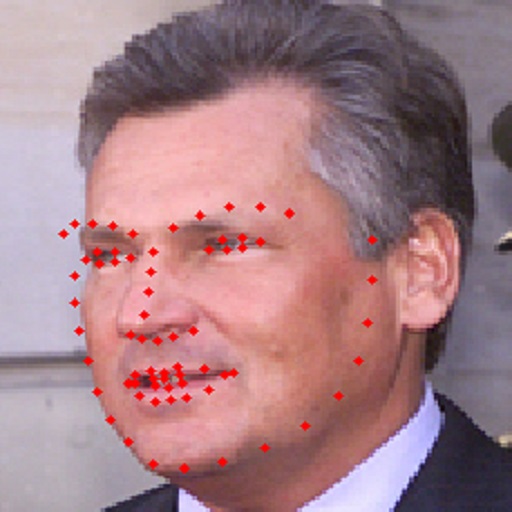}\hskip.2pt\includegraphics[width=2.2cm]{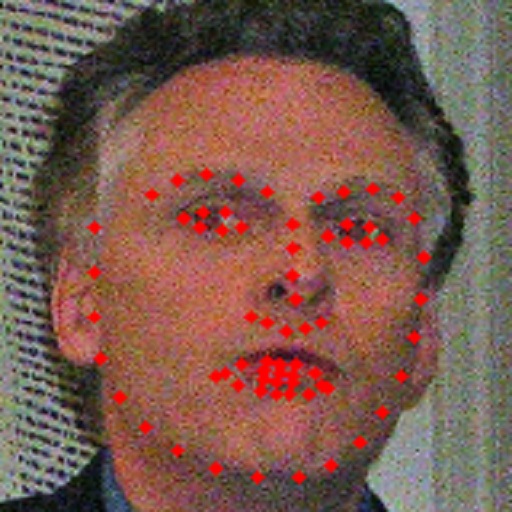}
 \hskip.2pt\includegraphics[width=2.2cm]{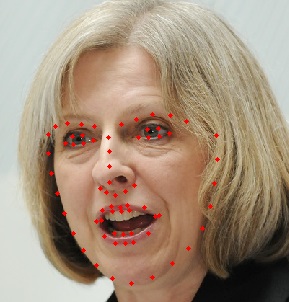}\hskip.2pt\includegraphics[width=2.2cm]{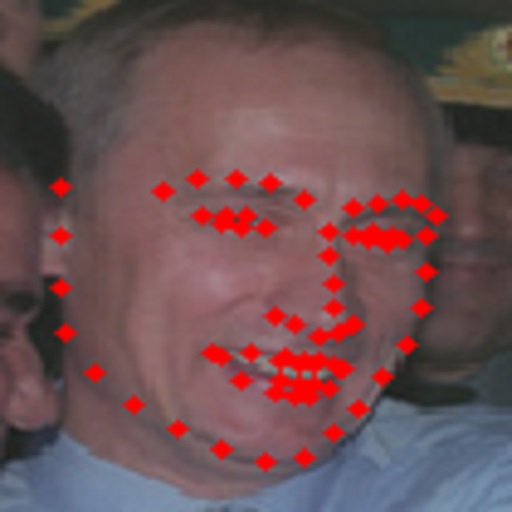}
 \vskip -2pt
\caption{ Qualitative results of our landmark localization method.  First row: LFPW, Second row: Helen, Third row: AFW and Fourth row: IBUG. Fifth row: IJB-A.}
\label{fig:landmark_images}
\end{figure*}

\begin{figure}[htp]
 \centering
\includegraphics[width=6.5cm]{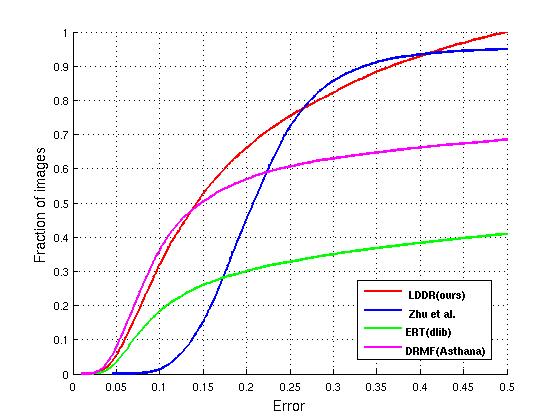}
\vskip 4pt
\caption{Average 3-pt error (normalized by eye-nose distance) vs fraction of images in the IJB-A dataset.}
\label{fig:Errors_IJBA}
\end{figure}

Tables~\ref{tbl:LFPW}, \ref{tbl:Helen}, \ref{tbl:ibug} and Figure \ref{fig:Errors} provide the Normalized Mean Square Error and average pt-pt error (normalized by face size) vs fraction of images plots of different methods, respectively.  In Figure~\ref{fig:Errors_IJBA}  we present the comparison of our algorithm with \cite{AFW_dataset_CVPR2012}, \cite{Asthana:2013:RDR:2514950.2516059} and \cite{kazemi2014one}.  Our deep descriptor-based global shape regression method outperforms the above mentioned state-of-the-art methods. The tables also show a comparison of our method with many other pioneering methods such as Gauss Newton based Deformable Part Models \cite{6909635} and Robust Cascaded Pose Regression (RPCR) \cite{10.1109/ICCV.2013.191} and some recent methods like \cite{Zhu_2015_CVPR}.  Figure~\ref{fig:landmark_images} shows some landmark localization results on the five datasets. It can be seen from this figure that our method is able to localize landmarks on near profile faces as well as faces of low resolution, partially visible and expression from the \textit{IJB-A} dataset.  

Randomly rotating and flipping doubles the amount of data and hence generalizes the data more while reducing the error by $\thicksim2$\%. After the advent of deep learning, it was seen that the $conv_5$ features capture a lot of salient information. Our method depends on the generalization of the \textit{deep descriptors} and hence the increase in the amount of data available for training favors the learning step. After training only on \textit{Helen} and LFPW trainset, we get an error of 5.09\% and 5.08\%, respectively.  However, after training on the cumulative data we achieve better performance getting 4.76\% on the former and 4.67\% on the latter.  Also, it can be seen from Tables~\ref{tbl:LFPW} and \ref{tbl:Helen}, the error in $68$ landmark points is higher than that in $49$ points as the former includes the face contour points. It is evident from our experiments that the proposed method performs better than \cite{AFW_dataset_CVPR2012} and \cite{XiongD13} where HOG and SIFT  were used as their features. Table \ref{tbl:ibug} shows the performance of our method on challenging subset of 300-W ibug dataset. The error in the performance of CFSS \cite{Zhu_2015_CVPR} is lower than our method.  This may be due to the fact that  CFSS performs its initial search on the space of multiple mean shapes, whereas we initialize with only one mean shape at test time. We do this to reduce the time and space complexity during training. In our experiments we only flipped and rotated the shapes in contrast to conventional techniques where the shapes are flipped, rotated, translated and scaled. This also demonstrates the discriminatory quality of our Deep Descriptors and how better it can get given a large amount of diversified training data.   

\begin{table}[thp!]
\begin{center}
\resizebox{\linewidth}{!}{%
\begin{tabular}{|p{4.1cm}|p{1.5cm}|p{1.5cm}|}
\hline
\centering {\it Method} & \centering {\it 68-pts} &  {\it 49-pts}\\
\hline\hline
\centering Zhu \it{et al.} \cite{AFW_dataset_CVPR2012}   & \centering 8.16         & 7.43  \\
\centering DRMF \cite{Asthana:2013:RDR:2514950.2516059}         &\centering 6.70         & -      \\
\centering RCPR \cite{10.1109/ICCV.2013.191}         &\centering 5.93         & 4.64  \\
\centering SDM \cite{XiongD13}          & \centering 5.50         & 4.25       \\
\centering GN-DPM \cite{6909635}       & \centering 5.69         & 4.06    \\
\centering CFAN \cite{CFAN}       & \centering 5.53         & -        \\
\centering CFSS \cite{Zhu_2015_CVPR}       & \centering 4.63         & 3.47        \\
\hline
\centering {\bf LDDR}   & \centering {\bf 4.76}   & {\bf 2.36}\\
\hline
\end{tabular}}
\vskip 4pt
\caption{Averaged error comparison of different methods on the Helen dataset.} 
\label{tbl:Helen}
\end{center}
\end{table}

\subsection{Runtime} All the experiments were performed using an NVIDIA TITAN-X GPU using cudnn library on a 2.3Ghz computer. Training on LFPW took 5.5 hours and on Helen took 9 hours. Training on cumulative data took around 15 hours. Due to different CNN being initialized in each stage, the testing was observed to be slow taking $\thicksim 4$ seconds given a face bounding box. \textit{However in our implementation testing was close to real time performance taking only $\thicksim 0.8$ seconds per face, hereby reducing the testing time by $80\% $ }. This includes the time taken for feature extraction and regression. The time consuming part for the landmark localization was the initialization of a different CNN in each stage. To counter this delay in testing, we merged the 4 CNN models in a single CNN model which is initialized only once. To reduce the performance time even more, the 68 patches extracted around the intermediate shape were passed in a batch. 

\begin{table}[thp!]
\begin{center}
\begin{tabular}{|p{7cm}|p{.9cm}|}
\hline
\centering {\it Method} &  {\it 68-pts} \\
\hline\hline
\centering Zhu \it{et al.} \cite{AFW_dataset_CVPR2012}   &18.33  \\
\centering DRMF \cite{Asthana:2013:RDR:2514950.2516059}         & 19.75    \\
\centering RCPR \cite{10.1109/ICCV.2013.191}         & 17.26  \\
\centering SDM \cite{XiongD13}          & 15.40   \\
\centering GN-DPM \cite{6909635}       &-  \\
\centering CFAN \cite{CFAN}       &-  \\
\centering ESR \cite{DBLP:journals/ijcv/CaoWWS14}               & 17.00 \\
\centering LBF \cite{DBLP:conf/cvpr/RenCW014}               & 11.98  \\
\centering LBF Fast \cite{DBLP:conf/cvpr/RenCW014}          & 15.50\\
\centering CFSS \cite{Zhu_2015_CVPR}          & 9.98\\
\hline
\centering {\bf LDDR}   &  {\bf 11.49}\\
\hline
\end{tabular}
\caption{Averaged error comparison of different methods on the iBUG  challenging dataset.} 
\label{tbl:ibug}
\end{center}
\end{table}

\section{Conclusions}\label{sec:conc}
In this paper, we presented a deep descriptor-based method for face alignment using regression of local descriptors.  
The highly informative nature of \textit{deep descriptor} makes it useful as SIFT, SURF and HOG features. This means \textit{deep descriptors} have potential in many different kinds of applications in machine vision, such as pose estimation, activity recognition and human detection and many others. We also presented an effective way of reducing the testing time by combining four CNNs into one achieving real-time performance.  Extensive experiments on five publicly available unconstrained face datasets demonstrate the effectiveness of our proposed image alignment approach.

\section*{Acknowledgments}
This research is based upon work supported by the Office of the Director of National Intelligence (ODNI), Intelligence Advanced Research Projects
Activity (IARPA), via IARPA R\&D Contract No. 2014-14071600012. The views and conclusions contained herein are those of the authors and should
not be interpreted as necessarily representing the official policies or endorsements, either expressed or implied, of the ODNI, IARPA, or the U.S. Government. The U.S. Government is authorized to reproduce and distribute reprints for Governmental purposes notwithstanding any copyright annotation
thereon.
{\small
\bibliographystyle{ieee}
\bibliography{FD}

\begin{thebibliography}{10}\itemsep=-1pt

\bibitem{Antonakos_2015_CVPR}
E.~Antonakos, J.~Alabort-i Medina, and S.~Zafeiriou.
\newblock Active pictorial structures.
\newblock June 2015.

\bibitem{Asthana:2013:RDR:2514950.2516059}
A.~Asthana, S.~Zafeiriou, S.~Cheng, and M.~Pantic.
\newblock Robust discriminative response map fitting with constrained local
  models.
\newblock In {\em Proceedings of the 2013 IEEE Conference on Computer Vision
  and Pattern Recognition}, CVPR '13, pages 3444--3451, Washington, DC, USA,
  2013. IEEE Computer Society.

\bibitem{Belhumeur:2011:LPF:2191740.2192193}
P.~N. Belhumeur, D.~W. Jacobs, D.~J. Kriegman, and N.~Kumar.
\newblock Localizing parts of faces using a consensus of exemplars.
\newblock In {\em Proceedings of the 2011 IEEE Conference on Computer Vision
  and Pattern Recognition}, CVPR '11, pages 545--552, Washington, DC, USA,
  2011. IEEE Computer Society.

\bibitem{10.1109/ICCV.2013.191}
X.~P. Burgos-Artizzu, P.~Perona, and P.~Dollar.
\newblock Robust face landmark estimation under occlusion.
\newblock {\em Computer Vision, IEEE International Conference on},
  0:1513--1520, 2013.

\bibitem{DBLP:journals/ijcv/CaoWWS14}
X.~Cao, Y.~Wei, F.~Wen, and J.~Sun.
\newblock Face alignment by explicit shape regression.
\newblock {\em International Journal of Computer Vision}, 107(2):177--190,
  2014.

\bibitem{AAM}
T.~Cootes, G.~Edwards, and C.~Taylor.
\newblock Active appearance models.
\newblock {\em Pattern Analysis and Machine Intelligence, IEEE Transactions
  on}, 23(6):681--685, Jun 2001.

\bibitem{Cootes:1995:ASM:206543.206547}
T.~F. Cootes, C.~J. Taylor, D.~H. Cooper, and J.~Graham.
\newblock Active shape models\&mdash;their training and application.
\newblock {\em Comput. Vis. Image Underst.}, 61(1):38--59, Jan. 1995.

\bibitem{Cristinacce06featuredetection}
D.~Cristinacce and T.~Cootes.
\newblock Feature detection and tracking with constrained local models.
\newblock pages 929--938, 2006.

\bibitem{REF08a}
R.-E. Fan, K.-W. Chang, C.-J. Hsieh, X.-R. Wang, and C.-J. Lin.
\newblock {LIBLINEAR}: A library for large linear classification.
\newblock {\em Journal of Machine Learning Research}, 9:1871--1874, 2008.

\bibitem{10.1109/TPAMI.2012.231}
C.~Farabet, C.~Couprie, L.~Najman, and Y.~LeCun.
\newblock Learning hierarchical features for scene labeling.
\newblock {\em IEEE Transactions on Pattern Analysis and Machine Intelligence},
  35(8):1915--1929, 2013.

\bibitem{girshick14CVPR}
R.~Girshick, J.~Donahue, T.~Darrell, and J.~Malik.
\newblock Rich feature hierarchies for accurate object detection and semantic
  segmentation.
\newblock In {\em Computer Vision and Pattern Recognition}, 2014.

\bibitem{Gross:2005:GVP:1709142.1709186}
R.~Gross, I.~Matthews, and S.~Baker.
\newblock Generic vs. person specific active appearance models.
\newblock {\em Image Vision Comput.}, 23(12):1080--1093, Nov. 2005.

\bibitem{Gross:2006:AAM:1709247.1709267}
R.~Gross, I.~Matthews, and S.~Baker.
\newblock Active appearance models with occlusion.
\newblock {\em Image Vision Comput.}, 24(6):593--604, June 2006.

\bibitem{Gross:2010:MUL:1746745.1747071}
R.~Gross, I.~Matthews, J.~Cohn, T.~Kanade, and S.~Baker.
\newblock Multi-pie.
\newblock {\em Image Vision Comput.}, 28(5):807--813, May 2010.

\bibitem{fddbTech}
V.~Jain and E.~Learned-Miller.
\newblock Fddb: A benchmark for face detection in unconstrained settings.
\newblock Technical Report UM-CS-2010-009, University of Massachusetts,
  Amherst, 2010.

\bibitem{jia2014caffe}
Y.~Jia, E.~Shelhamer, J.~Donahue, S.~Karayev, J.~Long, R.~Girshick,
  S.~Guadarrama, and T.~Darrell.
\newblock Caffe: Convolutional architecture for fast feature embedding.
\newblock {\em arXiv preprint arXiv:1408.5093}, 2014.

\bibitem{kazemi2014one}
V.~Kazemi and J.~Sullivan.
\newblock One millisecond face alignment with an ensemble of regression trees.
\newblock In {\em CVPR}, 2014.

\bibitem{F._2015_CVPR}
B.~F. Klare, B.~Klein, E.~Taborsky, A.~Blanton, J.~Cheney, K.~Allen,
  P.~Grother, A.~Mah, M.~Burge, and A.~K. Jain.
\newblock Pushing the frontiers of unconstrained face detection and
  recognition: Iarpa janus benchmark a.
\newblock June 2015.

\bibitem{NIPS2012_4824}
A.~Krizhevsky, I.~Sutskever, and G.~E. Hinton.
\newblock Imagenet classification with deep convolutional neural networks.
\newblock In F.~Pereira, C.~Burges, L.~Bottou, and K.~Weinberger, editors, {\em
  Advances in Neural Information Processing Systems 25}, pages 1097--1105.
  Curran Associates, Inc., 2012.

\bibitem{Le:2012:IFF:2403072.2403124}
V.~Le, J.~Brandt, Z.~Lin, L.~Bourdev, and T.~S. Huang.
\newblock Interactive facial feature localization.
\newblock In {\em Proceedings of the 12th European Conference on Computer
  Vision - Volume Part III}, ECCV'12, pages 679--692, Berlin, Heidelberg, 2012.
  Springer-Verlag.

\bibitem{Lee_2015_CVPR}
D.~Lee, H.~Park, and C.~D. Yoo.
\newblock Face alignment using cascade gaussian process regression trees.
\newblock June 2015.

\bibitem{conf/eccv/LiangXWS08}
L.~Liang, R.~Xiao, F.~Wen, and J.~S. 0001.
\newblock Face alignment via component-based discriminative search.
\newblock In D.~A. Forsyth, P.~H.~S. Torr, and A.~Zisserman, editors, {\em ECCV
  (2)}, volume 5303 of {\em Lecture Notes in Computer Science}, pages 72--85.
  Springer, 2008.

\bibitem{Lowe04distinctiveimage}
D.~G. Lowe.
\newblock Distinctive image features from scale-invariant keypoints.
\newblock {\em International Journal of Computer Vision}, 60:91--110, 2004.

\bibitem{Matthews:2004:AAM:993451.996344}
I.~Matthews and S.~Baker.
\newblock Active appearance models revisited.
\newblock {\em Int. J. Comput. Vision}, 60(2):135--164, Nov. 2004.

\bibitem{Messer99xm2vtsdb:the}
K.~Messer, J.~Matas, J.~Kittler, and K.~Jonsson.
\newblock Xm2vtsdb: The extended m2vts database.
\newblock In {\em In Second International Conference on Audio and Video-based
  Biometric Person Authentication}, pages 72--77, 1999.

\bibitem{DBLP:conf/cvpr/RenCW014}
S.~Ren, X.~Cao, Y.~Wei, and J.~Sun.
\newblock Face alignment at 3000 {FPS} via regressing local binary features.
\newblock In {\em 2014 {IEEE} Conference on Computer Vision and Pattern
  Recognition, {CVPR} 2014, Columbus, OH, USA, June 23-28, 2014}, pages
  1685--1692, 2014.

\bibitem{ILSVRC15}
O.~Russakovsky, J.~Deng, H.~Su, J.~Krause, S.~Satheesh, S.~Ma, Z.~Huang,
  A.~Karpathy, A.~Khosla, M.~Bernstein, A.~C. Berg, and L.~Fei-Fei.
\newblock {ImageNet Large Scale Visual Recognition Challenge}.
\newblock {\em International Journal of Computer Vision (IJCV)}, 2015.

\bibitem{6595977}
C.~Sagonas, G.~Tzimiropoulos, S.~Zafeiriou, and M.~Pantic.
\newblock A semi-automatic methodology for facial landmark annotation.
\newblock In {\em Computer Vision and Pattern Recognition Workshops (CVPRW),
  2013 IEEE Conference on}, pages 896--903, June 2013.

\bibitem{DBLP:conf/cvpr/Saragih11}
J.~Saragih.
\newblock Principal regression analysis.
\newblock In {\em The 24th {IEEE} Conference on Computer Vision and Pattern
  Recognition, {CVPR} 2011, Colorado Springs, CO, USA, 20-25 June 2011}, pages
  2881--2888, 2011.

\bibitem{conf/iccv/SaragihLC09}
J.~M. Saragih, S.~Lucey, and J.~F. Cohn.
\newblock Face alignment through subspace constrained mean-shifts.
\newblock In {\em ICCV}, pages 1034--1041. IEEE, 2009.

\bibitem{Saragih:2011:DMF:1937966.1938021}
J.~M. Saragih, S.~Lucey, and J.~F. Cohn.
\newblock Deformable model fitting by regularized landmark mean-shift.
\newblock {\em Int. J. Comput. Vision}, 91(2):200--215, Jan. 2011.

\bibitem{Sauer11accurateregression}
P.~Sauer, T.~Cootes, and C.~Taylor.
\newblock Accurate regression procedures for active appearance models.
\newblock In {\em In BMVC}, 2011.

\bibitem{sermanet-iclr-14}
P.~Sermanet, D.~Eigen, X.~Zhang, M.~Mathieu, R.~Fergus, and Y.~LeCun.
\newblock Overfeat: Integrated recognition, localization and detection using
  convolutional networks.
\newblock In {\em International Conference on Learning Representations (ICLR
  2014)}. CBLS, April 2014.

\bibitem{Sun:2013:DCN:2514950.2516090}
Y.~Sun, X.~Wang, and X.~Tang.
\newblock Deep convolutional network cascade for facial point detection.
\newblock In {\em Proceedings of the 2013 IEEE Conference on Computer Vision
  and Pattern Recognition}, CVPR '13, pages 3476--3483, Washington, DC, USA,
  2013. IEEE Computer Society.

\bibitem{Taylor:2010:CLS:1888212.1888225}
G.~W. Taylor, R.~Fergus, Y.~LeCun, and C.~Bregler.
\newblock Convolutional learning of spatio-temporal features.
\newblock In {\em Proceedings of the 11th European Conference on Computer
  Vision: Part VI}, ECCV'10, pages 140--153, Berlin, Heidelberg, 2010.
  Springer-Verlag.

\bibitem{doi:10.5244/C.24.91}
P.~Tresadern, P.~Sauer, and T.~Cootes.
\newblock Additive update predictors in active appearance models.
\newblock In {\em Proceedings of the British Machine Vision Conference}, pages
  91.1--91.12. BMVA Press, 2010.
\newblock doi:10.5244/C.24.91.

\bibitem{6909635}
G.~Tzimiropoulos and M.~Pantic.
\newblock Gauss-newton deformable part models for face alignment in-the-wild.
\newblock In {\em Computer Vision and Pattern Recognition (CVPR), 2014 IEEE
  Conference on}, pages 1851--1858, June 2014.

\bibitem{Welinder10p.:cascaded}
P.~Welinder and P.~Perona.
\newblock P.: Cascaded pose regression.
\newblock In {\em In: IEEE Conference on Computer Vision and Pattern
  Recognition}, 2010.

\bibitem{global2015xiong}
X.~Xiong and F.~D. la~Torre.
\newblock Global supervised descent method.
\newblock In {\em CVPR}, 2015.

\bibitem{XiongD13}
Xuehan-Xiong and F.~{De la Torre}.
\newblock Supervised descent method and its application to face alignment.
\newblock In {\em IEEE Conference on Computer Vision and Pattern Recognition
  (CVPR)}, 2013.

\bibitem{Yan:2013:LCM:2586110.2586274}
J.~Yan, Z.~Lei, D.~Yi, and S.~Z. Li.
\newblock Learn to combine multiple hypotheses for accurate face alignment.
\newblock In {\em Proceedings of the 2013 IEEE International Conference on
  Computer Vision Workshops}, ICCVW '13, pages 392--396, Washington, DC, USA,
  2013. IEEE Computer Society.

\bibitem{CFAN}
J.~Zhang, S.~Shan, M.~Kan, and X.~Chen.
\newblock Coarse-to-fine auto-encoder networks (cfan) for real-time face
  alignment.
\newblock In D.~Fleet, T.~Pajdla, B.~Schiele, and T.~Tuytelaars, editors, {\em
  Computer Vision � ECCV 2014}, volume 8690 of {\em Lecture Notes in Computer
  Science}, pages 1--16. Springer International Publishing, 2014.

\bibitem{Zhu_2015_CVPR}
S.~Zhu, C.~Li, C.~Change~Loy, and X.~Tang.
\newblock Face alignment by coarse-to-fine shape searching.
\newblock June 2015.

\bibitem{AFW_dataset_CVPR2012}
X.~Zhu and D.~Ramanan.
\newblock Face detection, pose estimation, and landmark localization in the
  wild.
\newblock In {\em IEEE Conference on Computer Vision and Pattern Recognition},
  pages 2879--2886, June 2012.

\end{thebibliography}
}
\end{document}